\documentclass[letterpaper, 10 pt, conference]{ieeeconf}  

\IEEEoverridecommandlockouts                              

\usepackage{graphics} 
\usepackage{epsfig} 
\usepackage{mathptmx} 
\usepackage{times} 
\usepackage{amsmath} 
\usepackage{amssymb}  
\usepackage{algorithm}
\usepackage{algpseudocode}
\usepackage{epstopdf}
\usepackage{colortbl}
\usepackage{booktabs}
\usepackage{array}
\usepackage{etex}
\usepackage{adjustbox}
\usepackage{graphicx}
\usepackage{adjustbox,lipsum}
\usepackage[table]{xcolor}
\usepackage{amsmath}
\usepackage{mdwtab}
\usepackage{multirow}
\usepackage{colortbl}
\usepackage{url}
\usepackage{bm}
\usepackage{hyperref}
\usepackage{cleveref}

\title{\LARGE \bf
Social Navigation Planning Based on People's Awareness of Robots
}

\author{Minkyu Kim$^{1}$, Jaemin Lee$^{1}$, Steven Jens Jorgensen$^{1,2}$, and Luis Sentis$^{3}$
\thanks{$^{1}$M. Kim, J. Lee, and S. J. Jorgensen are with the Department of Mechanical Engineering, University of Texas at Austin, TX 78712, USA
        {\tt\small \{steveminq,jmlee87,stevenjj\}@utexas.edu }}%
\thanks{$^{2}$Supported by a NASA Space Technology Research Fellowship}%
\thanks{$^{3}$L. Sentis is with the Department of Aerospace Engineering and Engineering Mechanics, University of Texas at Austin,
         TX 78712, USA
        {\tt\small lsentis@austin.utexas.edu}}%
}

\begin{document}
\maketitle
\thispagestyle{empty}
\pagestyle{empty}

\begin{abstract}

When mobile robots maneuver near people, they run the risk of rudely blocking their paths—but not all people behave the same around robots. People that have not noticed the
robot are the most difficult to predict. This paper investigates how mobile robots can generate acceptable paths in dynamic environments by predicting human behavior. Here, human be-
havior may include both physical and mental behavior, we focus on the latter. We introduce a simple safe interaction model: when a human seems unaware of the robot, it should avoid
going too close. In this study, people around robots are detected and tracked using sensor fusion and filtering techniques. To handle uncertainties in the dynamic environment, a Partially-Observable Markov Decision Process Model (POMDP) is used to formulate a navigation planning problem in the shared environment. People’s awareness of robots is inferred and
included as a state and reward model in the POMDP. The proposed planner enables a robot to change its navigation plan based on its perception of each person’s robot-awareness. As far
as we can tell, this is a new capability. We conduct simulation and experiments using the Toyota Human Support Robot (HSR) to validate our approach. We demonstrate that the proposed
framework is capable of running in real-time.

\end{abstract}

\section{INTRODUCTION}

 In the near future, robots will interact, explore, and cooperate with human beings, and they will permeate into people's lives and our societies. This application calls for new social functions in robotic systems. Several studies have explored social navigation based on robots being aware of people \cite{sisbot2007human}\cite{nonaka2004evaluation} \cite{shi2008human}. Several studies introduced the concept of social distance, or proxemics, as a personal space that influences robot navigation performance \cite{mumm2011human} \cite{rios2015proxemics}. These kinds of studies highlighted the need to consider the social and cognitive side of people for effective navigation. However, they do not consider the people's awareness of robot which we propose as a new metric for robot navigation.

 To insert human parameters (social parameters) into planning problems, a human-aware motion planner has been used to not only provide safe robot paths, but also to synthesize socially-acceptable and legible paths in the presence of humans \cite{sisbot2007human}. Another study characterizes the concept of comfort, naturalness, and sociability of navigation performance \cite{kruse2013human}. A popular method has considered the above social characteristics to build a cost map in combination with a conventional sampling-based planner \cite{mainprice2011planning}. On the other hand, there exist other approaches that utilize human teaching to navigate by demonstrating or providing feedback \cite{chernova2009interactive}. Although these kinds of researches showed us the possibility  of social navigation, real-time motion planning in human-crowded environments still has
not reached a satisfactory level, especially when navigating in crowds.

\begin{figure}[t]
    \centering
        {\includegraphics[width=1.0\linewidth]{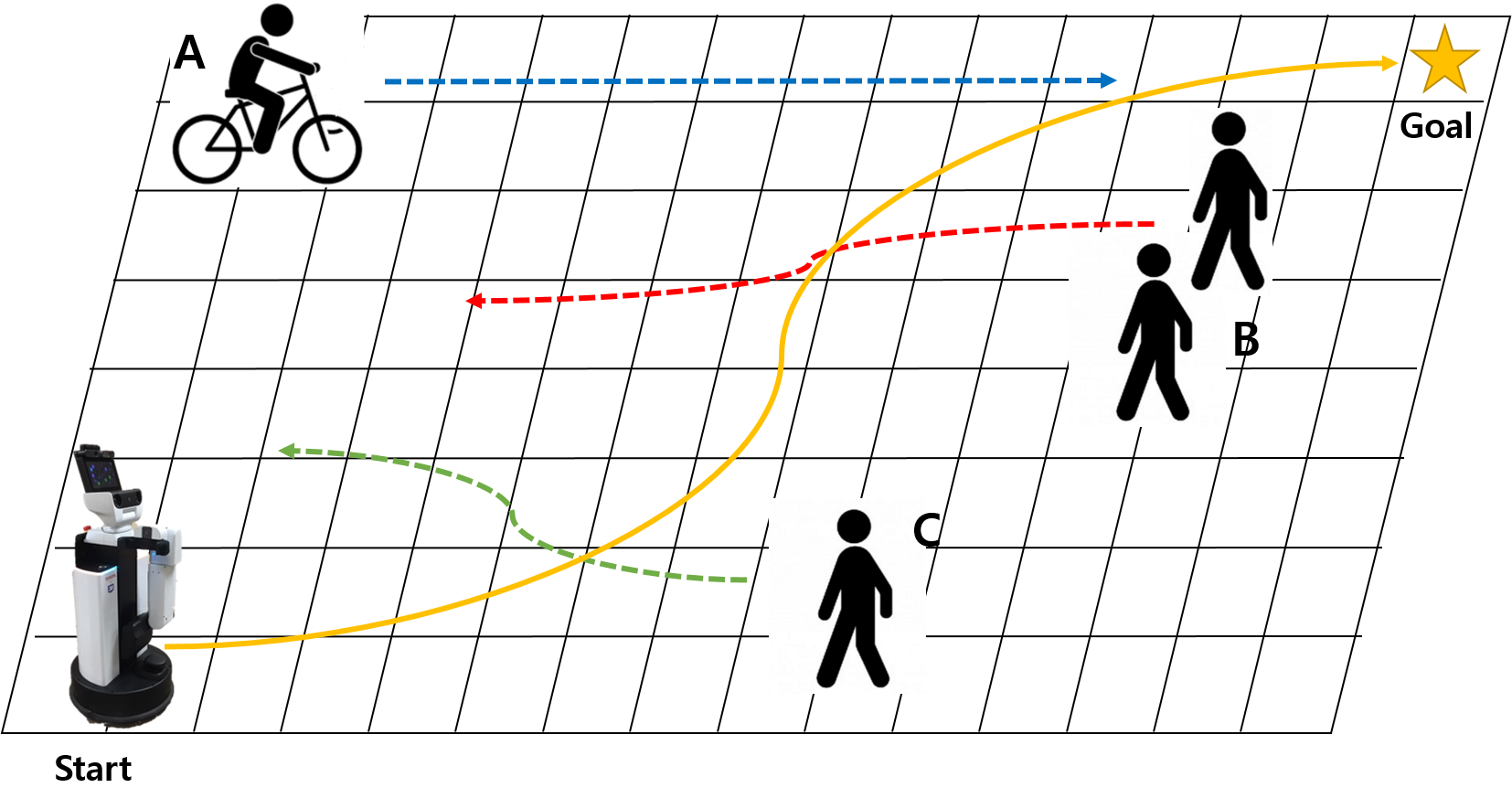}}
    \caption{Ambiguous situation in navigation: Who has the lowest trust to the robot? A. A person who is riding a bike with a high velocity? B. A couple who are talking to each other? C. A person who is quite close to the robot? Through incorporating notions of awareness, a robot can navigate confidently for each situation while maintaining a level of comfort for the surrounding people.}
       \label{Figure:Problem definition}
\end{figure}

For planning motions of mobile robots in dynamic situations, many studies regarding the prediction of pedestrians have been conducted, from a social force model \cite{helbing1995social} to machine learning-based estimations \cite{scovanner2009learning} \cite{keller2014will}. However, they did not consider an human's awareness state, which could be major factor for human robot interactions. Although the study presented in \cite{nonaka2004evaluation} uses mental states, it does not use them for navigation.

In fact, a human mental state is quite difficult to examine becauseit is fundamentally uncertain, both due to the unobservable and unpredictable nature of mental state; the same state could produce many different actions, while many different mental states could lead to the same action. Rather, this ambiguous problem can be simplified if it is confined to the problem of navigation with assumptions. We assume that a mental state can be only determined
by eye-contact between two agents. This assumption can be justified by studies from the social cognitive field \cite{macrae2002you} \cite{baker2014modeling} \cite{hollands2002look}. According to these studies, it can be confirmed that eye contact plays an important role in anticipating other's intentions for navigation. Focusing on this point, the proposed idea is quite simple: insert this eye contact model for robots to predict future movement of humans by defining
a mental state with the concept of awareness.

Many researchers have already pointed out that probabilistic representations of target states and reasoning over them are quite effective for navigation of mobile robots in dynamic environments. In situations of probabilistic decision making, Partially Observable Markov Decision Processes (POMDPs) have been widely used in robot navigation and human interactions \cite{foka2007real}. Since finding optimal control strategies in POMDP cases is computationally intractable due to the continuous and high dimensional belief space, POMDPs have usually been applied to topological navigation \cite{pineau2003point}.

In this paper, we propose a navigation planning framework of mobile robots based on human detection, POMDP and human awareness estimated by an eye contact (gaze) model.
More specifically, the proposed approach can be applied to human-crowded environments, including moving pedestrians and dynamic obstacles. The main contribution of this study
is to integrate human state estimation from real-time human detection and tracking and navigation planning to manage uncertainties, both position and awareness, from people. In
particular, the concept of human awareness of a robot is incorporated in the state model and the reward model of POMDP to improve social navigation performance in a way inspired by humans.

This paper is organized as follows. In section II, we provide background for the proposed methods containing Markov Decision Process (MDP) and POMDP models. Section III presents the framework of the proposed model, including detailed algorithms and the method for measuring human awareness based on gaze detection. In Section IV, the proposed methodology is validated in simulation and real hardware. We discuss the simulation results in terms of the effect of awareness in the navigation process. 

\section{Background}

\subsection{MDP $\&$ POMDP}

The basic concept of MDP is that of a decision-making problem formulated as a set of states and actions given defined costs. A crucial assumption in MDP is that the transition of states is a Markov process, and that their future distribution is conditionally independent of the history of states and only affected by the current state. Highly probable states are determined by a reward (so-called "value") function. The goal of the agent is to select an action which will generate the maximum value for a predetermined time horizon. 

Partially Observable MDP, or POMDP, is proposed to enable the MDP to be applied to the real world, which has a lot of uncertainties and disturbances. States usually cannot be measured directly, so we have to use incomplete sensors to perceive an environment. A POMDP model can be described by the tuple ($S, \pi, A, T, Z, E, R$), a finite set of states $S=\{s_1,\cdots s_{|S|}\} $, an initial probability distribution over these states $\pi$, a finite set of actions $A=\{a_1,\cdots a_{|A|}\} $, a finite set of observations $Z=\{z_1,\cdots z_{|Z|}\} $, and a transition function $T^{a,z}(s_i,s_j)=P(s_j|s_i,a,z)$ that maps $S\times A$ into discrete probability distributions over $S$. 

The transition model $T(s',a,s)$ specifies the conditional probability distribution of shifting from state $s$ to $s'$ by applying action policy $a$. $O(s',a,z)$ is the observation mapping that computes the probability of observing $z$ in state $s'$ when executing action $a$. Usually, the transition model and observation model can be rewritten as $T(s',a,s)= p(s'|s,a)$ , $Z(s',a,z) = P(z|s',a)$
where $s \in S$ , $a \in A$ , $z \in Z$.

\subsection{DESPOT}
 An on-line approach to solve POMDP is to combine planning and execution together through calculating and executing optimal action based on current belief state which is updated recursively over time. These on-line methods apply algorithmic techniques for computational efficiency. For example, heuristic search, branch-and-bound pruning, Monte Carlo sampling \cite{silver2010monte}, POMCP \cite{ross2008online}, and DESPOT \cite{somani2013despot} are among the fastest on-line POMDP methods recently.

 We adopted DESPOT as an on-line POMDP solver. The key concept of DESPOT is to reduce all policies under $K$ sampled scenarios. Under each scenario, a policy traces out a path in the belief tree consisting of a particular sequence of actions and observations. DESPOT is defined by a tree, which keeps only the belief-tree nodes and edges that are generated from all policies under the sampled scenarios. Assuming that there is height, $H$, in the belief tree, DESPOT is more sparse, only including $O(|A|^HK)$ nodes, than the original belief tree, which contains $O(|A|^H|Z|^HK)$ nodes, leading to a dramatic improvement in computational efficiency for moderate $K$ values. Equally importantly, it can be proven that a small DESPOT tree is sufficient to generate suboptimal policy, which admits a compact representation, with bounded regret \cite{somani2013despot}.

 \begin{figure}[b]
    \centering
       {\includegraphics[width=0.98\linewidth]{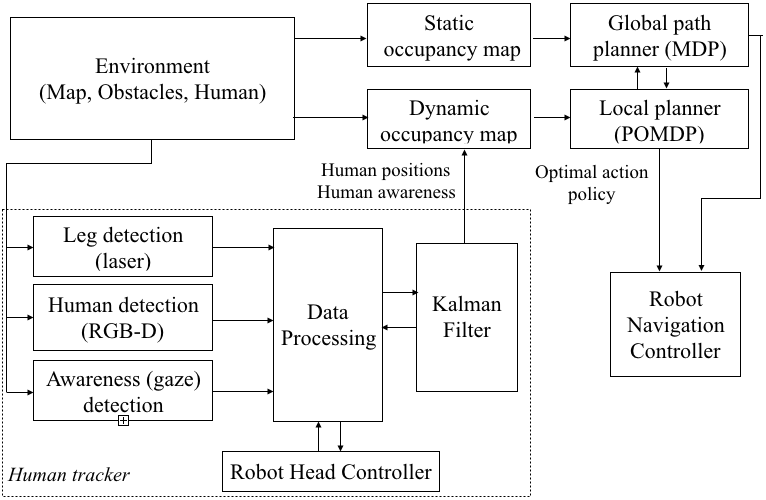}}
    \caption{Overall flowchart of proposed framework : a global planner calculates a global path while local planner seeks the optimal action policy. States for local planner (POMDP) are determined based on the combination of the human position and awareness from Human tracker (dotted block) and dynamic occupancy grid. Finally, a global path and an optimal action policy are delivered to the robot navigation controller.}
       \label{Figure:OverallFramework}
\end{figure}

\section{Methods}
\subsection{Overall framework}
 Our framework consists of double-layered planners: a global path planner with MDP and a local planner with POMDP, as shown in Fig.\ref{Figure:OverallFramework}. A global planner generates a collision-free path based on the environment, which can be pre-known or perceived in advance. This collision-free path can be described as a static occupancy grid map, which considers only static obstacles. On the other hand, a local planner acts as a reactive planner to cope with various possible situations that can occur during real-time navigation. Human tracking continuously detects and tracks people in a local window, a relatively small area around the robot, in order to react according to the movement of people. A POMDP-based local planner is designed to search for the optimal action policy that can obtain the maximum reward based on the belief states which can be estimated from observations. The proposed framework properly discretizes the global path given bound on the sub-path segments until arriving at the final goal position. Based on the action policy calculated from the local planner, the robot navigation controller finds the input command for the robot to reach the desired position.

\begin{algorithm}[t]
\caption{MDP Planner ($\mathcal{M}$)}
\begin{algorithmic}
\Procedure {MDPsolve}{}
\renewcommand{\algorithmicrequire}{\textbf{Input:}}
\renewcommand{\algorithmicensure}{\textbf{Output:}}
 \Require {$S$$\,$(state), $A$$\,$(Action), $R$$\,$(Reward), $T$$\,$(Transition)}
    \Ensure{$\pi^*(s)$ : The Optimal action policy} 
\State $V_0(s) \leftarrow 0$, $\,$ $k \leftarrow 0$
\Repeat
\For {$k \leftarrow k+1$}
\ForAll {$s \in S$} \\ 
    $V_k(b) =\max_{a \in A} \{\sum_{s^{'}}T(s,a,s^{'})[R(s,a,s^{'})+\gamma V_{k-1}(s^{'})] $  \EndFor
\EndFor
\Until{$\forall$ $s\|V_k(s)-V_{k-1}(s) \| < \epsilon $}
\ForAll {$s \in S$} \\ 
$\pi^{*}(s)=argmax_{a} \{\sum_{s^{'}}T(s,a,s^{'})[R(s,a,s^{'})+\gamma V_{k}(s^{'})]$
\EndFor
\EndProcedure
\end{algorithmic}
\end{algorithm}

\subsection{Global Path Planning: MDP}
\begin{figure}[t]
   \centering
    \includegraphics[width = 0.88\linewidth]{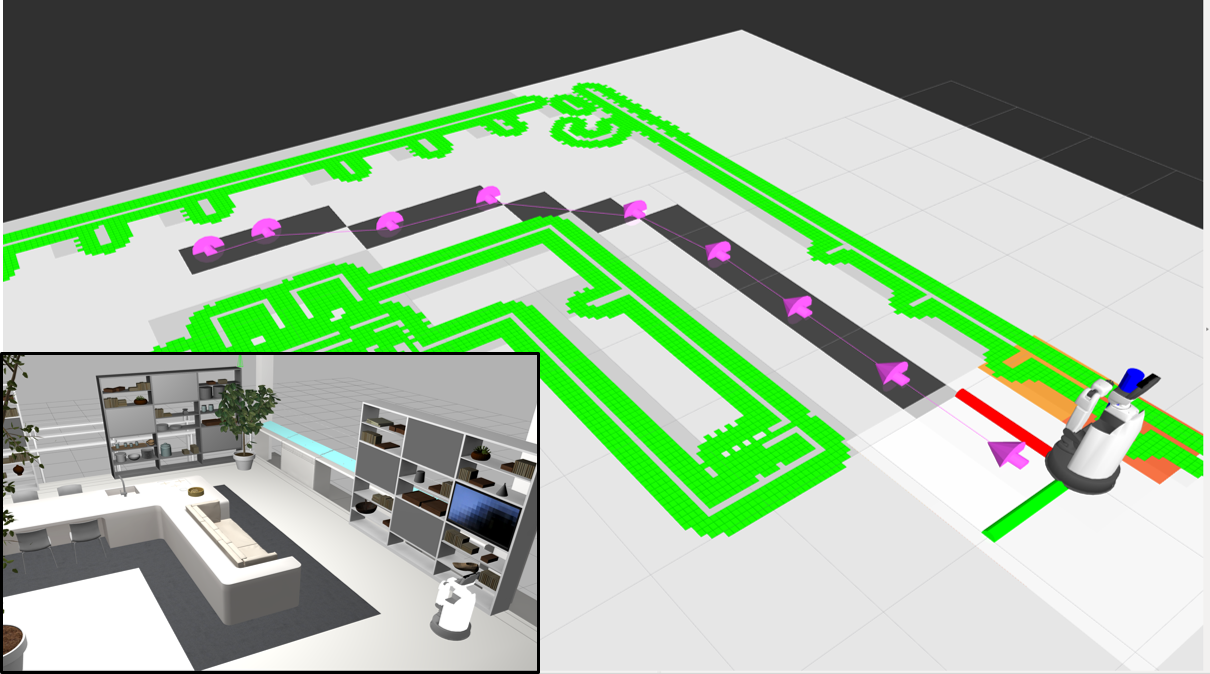}
 \label{Figure:MDP_path}
  \caption{ Global path (MDP solution) in a grid space, and the robot view.}
\end{figure}

The 2D environment can be represented by an occupancy grid, which is described as $o(i,j)$ for $i$, $j$ are 2D coordinates of grid map respectively \cite{elfes1989using}. This occupancy value becomes 0 if it is free while it becomes 1 or 2 if it is occupied with obstacles or human. Assuming that there is no human and mapping and localization are done in advance, based on this grid map occupancy, MDP-based global planner finds a collision-free path as shown in \ref{Figure:MDP_path}. Start position is the current robot position and goal position can be set with Graphic User Interface. In fact, this MDP planner obtains mapping from occupancy grid to action policy ($\mathcal{M}:\Re^1 \rightarrow \Re^1$). Value iteration can be used to solve the MDP problem as shown in Algorithm 1. Since we have optimal solutions for all the lattice points (grid map), robot is able to flexibly cope with the environment. Then, every cell which is not occupied with obstacles can have a desired action policy:

 \begin{equation}
      \bar{S}(i,j)=
  \Big\{ \begin{array}{cl} \pi^*(s) \in A & \textrm{if} \quad o(i,j) = 0 \quad (Free) \\ \phi & \textrm{otherwise} \quad (Occupied) \end{array}
\end{equation}
where $A = \{E,EN,N,NW,W,WS,S,SE\}$, or eight possible action sets, and $o(i,j)$ means occupancy grid. Particularly, starting from the robot's current position, a desired path can be generated by choosing consecutive grid cells until reaching goal position. Once the robot reaches the goal position, this path can be represented as 

\begin{equation}
	\bm{P} = \left\{ \left(p^1_x, p^1_y \right), \left(p^2_x, p^2_y \right)\cdots, \left(p^n_x, p^n_y \right) \right\}_{ \mathcal{D}_{grid}}
\end{equation}

where $\mathcal{D}_{grid}$ is the resolution of grid, which is set as 0.75 (m) in this study. We selected this value so that actual mobile robot hardware fits in one; 0.75 (m) $\times$ 0.75 (m) cell, which is applicable and reasonable for real hardware implementation. Based on this path information, we construct a smooth path with cubic spline interpolation methods. Fig. \ref{Figure:MDP_path} shows the collision-free path with static obstacles. 
  
\subsection{People Detection \& Tracking}
In order to successfully navigate human-robot coexisting environment, real-time detection and tracking of people are essential. In practice, tracking a human is quite a challenging task due to the limitation of sensor visibility, noise in sensor readings, possible target occlusions, and confusions from multiple targets. (RGB-D camera and laser scan data) is used to detect and track humans around the robot. This sensor fusion technique can improve the accuracy of detection and tracking people.
 
 \subsubsection{Vision-based Detection} 
\begin{figure}[t]
    \centering
        {\includegraphics[width=1.0\linewidth]{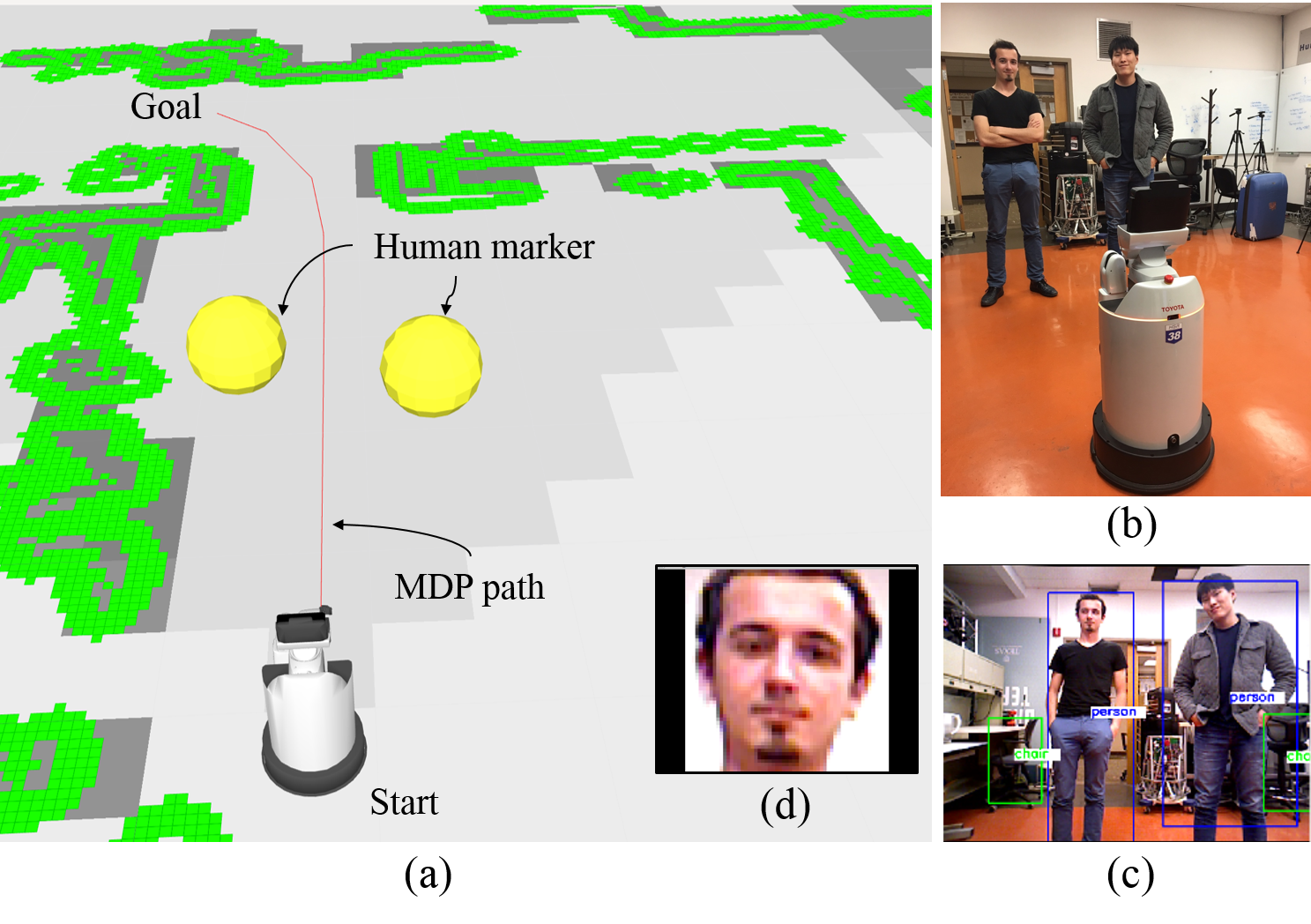}}
       \caption{ (a) Robot in grid space with the global path and human in front of it. (b) Snapshot of real laboratory. (c) Image-based human detection with YOLO. (d) Cropped face from face recognition.}
       \label{Figure:human_recognition}
\end{figure}

 For the vision based detection algorithm, one of the state of the art, a deep learning based real-time object detection algorithm, or You Only Look Once (YOLO) \cite{redmon2016you} is implemented to detect the existence of people. This algorithm has the advantage of being able to accurately recognize human only with some part of the shape of a person. In addition, this method is known for its rapid detection capability, as time latency is less than 25 milliseconds with optimal settings, which is far faster than conventional vision-based approaches. We can obtain the number of existing humans in the camera region and approximate 3D position of findings. Using point cloud data from the RGB-D camera, 3D bounding boxes of the human can be obtained, which is described in Fig. \ref{Figure:human_recognition}.

\subsubsection {Laser Sensor-based Detection}
 Laser scan data also can be utilized to detect a human leg through pattern recognition. Human leg pattern can be detected as three categories - Leg Apart (LA), Forward Straddle (FS), and Single Leg (SL) with feature-based classifiers, as described in \cite{bellotto2009multisensor}. We adopted their work to find all possible human leg patterns based on laser scan information. An advantage of using laser sensors is that algorithm is able to rapidly detect human legs when people are moving at a high speed, while vision-based methods might not able to detect human movements due to their processing and computational costs.

 \subsubsection{Measurement Fusion- Data processing}
  The main reason for using sensor fusion method is to simultaneously utilize the advantages of the observations. To combine two different types of observations (vision-based and laser-based measurements), the measurement fusion method is known for its effectiveness to produce better estimation results. We assumed that YOLO detection is much more reliable because the laser detection algorithm only provides possible candidates of human leg, rather than exact human leg information. This means that laser-based observations need to be filtered based on YOLO detections. Consequently, once robot detects the number of humans around it, referring to $N$, data processing method is designed to extract the same number of human legs out of all candidates from laser sensors. This processing method can be used in combination with gaze control since two sensor's field of view are  different. Outputs of data processing are delivered to the Kalman filter.
  
 \subsubsection{Kalman filter}
 The position of each detected humans is individually tracked over time using the Kalman filter technique \cite{julier1997new}. The Kalman filter algorithm consists of two steps: prediction and correction (update step). The first step predicts the current state from the previous states and the second step uses sensor measurement to update or correct the estimation from the previous step.   
 Each filter estimates each position of human candidates, or $x_k$, over time as one element of target clusters which is denoted as $X_k=\{x^1_k,x^2_k, \cdots x_k^N\}$, where N is the total target numbers at time step $k$. Here, a state includes 2D position and velocity of particle as $x_k^i=(x,y,\dot{x}, \dot{y})^T$, and the Kalman filter model uses the linear dynamic model and measurement model which are formulated as 
 \begin{eqnarray}
 \dot{x}&=&Ax+Bu+w \\
  z&=&Hx+v
 \end{eqnarray}
 where $A$ is a state transition matrix, $B$ is the input matrix, $u$ is input variable, and $w$ is a white Gaussian noise with covariance $Q$. The measurement variable, $z$, can be modeled with $H$, which is the observation matrix, and $v$ is observation noise variable, of which the covariance is $R$. Based on this formulation, the Kalman filter iteratively estimates a state variable, $x$, with consequent measurements, $z$. System model parameters used in this study can be referred to in the previous study \cite{leigh2015person}.
 
 \subsubsection{Face Recognition}
 \label{face_recognition}
  The face recognition package from \cite{ageitgey2013} is also implemented in our system. It basically uses Histogram of Oriented Gradients (HOG) \cite{dalal2005histograms} to detect faces and face landmark estimation \cite{kazemi2014one} to extract face features. Then, extracted features are used to train Deep Convolutional Neural Network to recognize face. Therefore, human faces can be recognized from the video stream, which leads to capturing human gaze. This face recognition is only used for gaze detection, not for human detection and tracking.
  
\subsubsection{Gaze detection (Awareness)}
\begin{figure}[t]
    \centering
        {\includegraphics[width=1.0\linewidth]{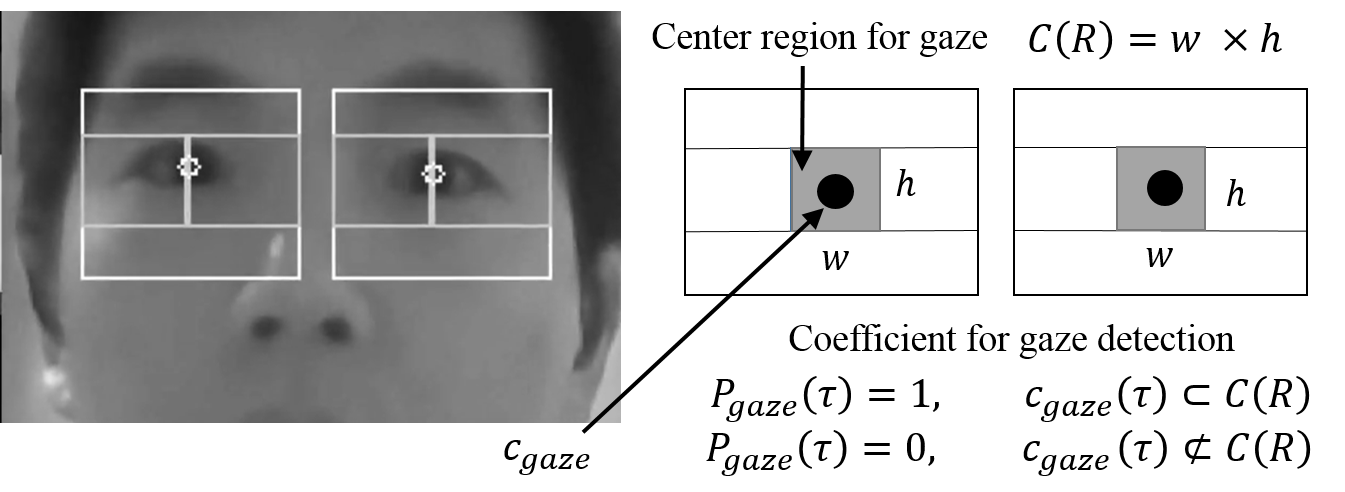}}
 		\caption{Gaze detection model}
       \label{Figure:gaze detection model description}
\end{figure}

 In order to observe $p_{gaze}$, given face image from the above algorithm, a gaze tracker using a simple image gradient-based eye center algorithm is applied to track the gaze of a human \cite{timm2011accurate}. This algorithm provides relatively high accuracy results with low computational cost.
 The robot's success of eye contact recognition is proportional to the time a person stares at the robot. It can be detected based on the fact that the human eyes are looking at the center of the camera of the robot. Therefore, the proposed method calculates time spent with both eyes on the center rectangular region, which can be defined as $C(R)$ with width $w$, and height $h$, as shown in Fig.\ref{Figure:gaze detection model description}. If measured time duration exceeds the time threshold, or $\epsilon$, then, the awareness variable is activated. This time threshold can be determined as 5 seconds through a trial and error approach. Based on this detection model, we can judge whether a person is aware of a robot or not, which can be described as the awareness variable. Consequently, the following equation is used to define the awareness variable
\begin{equation}
      G(t)=
  \Big\{ \begin{array}{cc} 1 & \int_{t_0}^{t}{P_{gaze}}(\tau)d\tau> \epsilon \\ -1 & \textrm{otherwise} \end{array}
\end{equation}

 If a person is aware of a robot, when the awareness variable is active, the robot is allowed to approach closer than to those who did not make eye contact. This feature can be obtained by differentiating the reward model according to pedestrians. In our model, the collision reward is individually set for each observed pedestrian to change the permission range of distance between human and robot.   
 
 \subsection{Local planner: POMDP}
 \subsubsection{States}
  As a local planner, we utilize the on-line POMDP solver, DESPOT, to obtain the optimal action policy based on current belief. The state variable contains a robot state and people state. A robot state includes robot position, $R_{pos}$, which can be measured directly by sensors in a local window describing the surroundings in the form of an occupancy grid. Pedestrians state contains the current position, and awareness, which are represented as $p_{pos}$ and $p_{awareness}$ respectively. The dimension of states varies according to the number of pedestrians in the local planner, $N$. To sum up, the state of the POMDP model can be described as
\begin{equation}
\begin{split}
\bm{S}  &=\{S_{\textrm{robot}}, \, S^1_{\textrm{ped}}, \, \cdots, \, S^N_{\textrm{ped}} \}. \\ 
S_{\textrm{robot}} &= [R_{pos}] \\
S_{\textrm{ped}} &=[p_{pos}, \, p_{awareness} ] 
\end{split}
\end{equation}
\label{Eq : POMDP STATE MODEL}
 
 \subsubsection{Observations}
 The observation information can be measured by a scanning laser range finder and a RGB-D Camera. The field of view of laser sensor is 270˚ range.  The distance range is assumed to be 5 meters ahead, to sense humans, static and dynamic obstacles. From this sensor model, our observation model is written as:


  In a simulation environment, each sensor has its own noise model, which is based on its specification. Based on sensor information, we can update our belief over the state variable at each time step.

\subsubsection{Actions (Policy Set)}
   Simplifying current mobile robot movements, we have 3 possible action policies: Go, Wait and Avoid. The "Go" policy make a robot go forward along the global path, while "Wait" means do not move for one time step. These two policies are included in the POMDP action set. Lastly, the "Avoid" action makes the robot move to the collision-free position when a robot can not move along the path due to pedestrians or dynamic obstacles. This action is only activated when the robot can not move for the predefined amount of time, and this command makes the global planner regenerate a global path, depending on dynamic occupancy grid. The default step size of this action policy is set to one grid resolution. Reducing the dimension of action policy sets can lessen the computational burden of the DESPOT algorithm.

  \begin{equation}
    \textrm{A}  = \{\textrm{Go},\, \textrm{Wait}, \, \textrm{Avoid}\} 
  \end{equation}

\begin{algorithm}[t]
\caption{Proposed Planner (Main loop)}
\begin{algorithmic}[1]
\State SetStartGoal()
\State $P\leftarrow$ MDPSolve(), $k \leftarrow 0$, $N\leftarrow \text{size}(P)$
\While{Not Goal} 
\State StateUpdate()  \Comment{Static/Dynamic Obs}
 \Repeat
\For {$k \leftarrow k+1$} 
 \State TrackerUpdate()  \Comment{People}
  \State BeliefUpdate() \Comment{Particle Filtering}
\State Action$^*$ $\leftarrow$ POMDPSolve()  \Comment{DESPOT}
\If {Action$^*$ $=$'Avoid'} 
    \State Go To Line 2
\EndIf
\State RobotControl(Action$^*$)   
\EndFor
\Until{$k<N$}
\EndWhile
\end{algorithmic}
\end{algorithm}

\subsubsection{Rewards}
 Establishing a reward model is quite a sensitive problem because we can design the characteristics of the desired action. Our proposed reward basically consists of reward for goal, penalty for collision, and time. Reward for goal state is set to the highest value. If a robot collides with a human or wall, reward gives a penalty. Lastly, since a navigation time is also one indicator that can evaluate performance, it is regarded as a penalty. Thus, reward function is written as 
  \begin{equation}
 R(s) = w_gR_{goal}(s)+w_cR_{col}(s)+w_tR_{time}(s)
  \end{equation}

 where $w_g$,$w_c$,and $w_g$ are weighting factors of each reward function. In particular, a human-collision reward model must be designed more carefully, depending on the degree of awareness that we defined above. A potential field approach \cite{ge2000new} is used to model the collision function based on the awareness variable and the distance between robot and human. This collision reward enables the planner to consider an awareness effect for each pedestrian. In other words, the reward collision model varies depending on awareness to change an acceptable permission range in equation (\ref{equation:R_col}). By differentiating this range, robot flexibly navigates with pedestrians who have different levels of awareness between robot and human. $R_{\textrm{Aware}}$ is set to smaller than $R_{\textrm{Non-Aware}}$ to allow the robot to have a high proximity to humans who are aware of it. 
      
\begin{equation}
      R_{col}=
  \Big\{ \begin{array}{cl} R_{col} & dist(R_{pos},P_{pos}) \leq \rho_{Aware} \\ 0 & \textrm{otherwise}  \end{array} 
  \label{equation:R_col}
\end{equation}

 \begin{equation}
   \rho_{Aware}=(\frac{1-G(k)}{2}){R_{\textrm{Aware}}}+(\frac{1+G(k)}{2}){R_{\textrm{Non-Aware}}} \\
  \label{Equation: Reward+collision}
\end{equation}

\begin{figure*}[t]
    \centering
        {\includegraphics[width=1.0\linewidth]{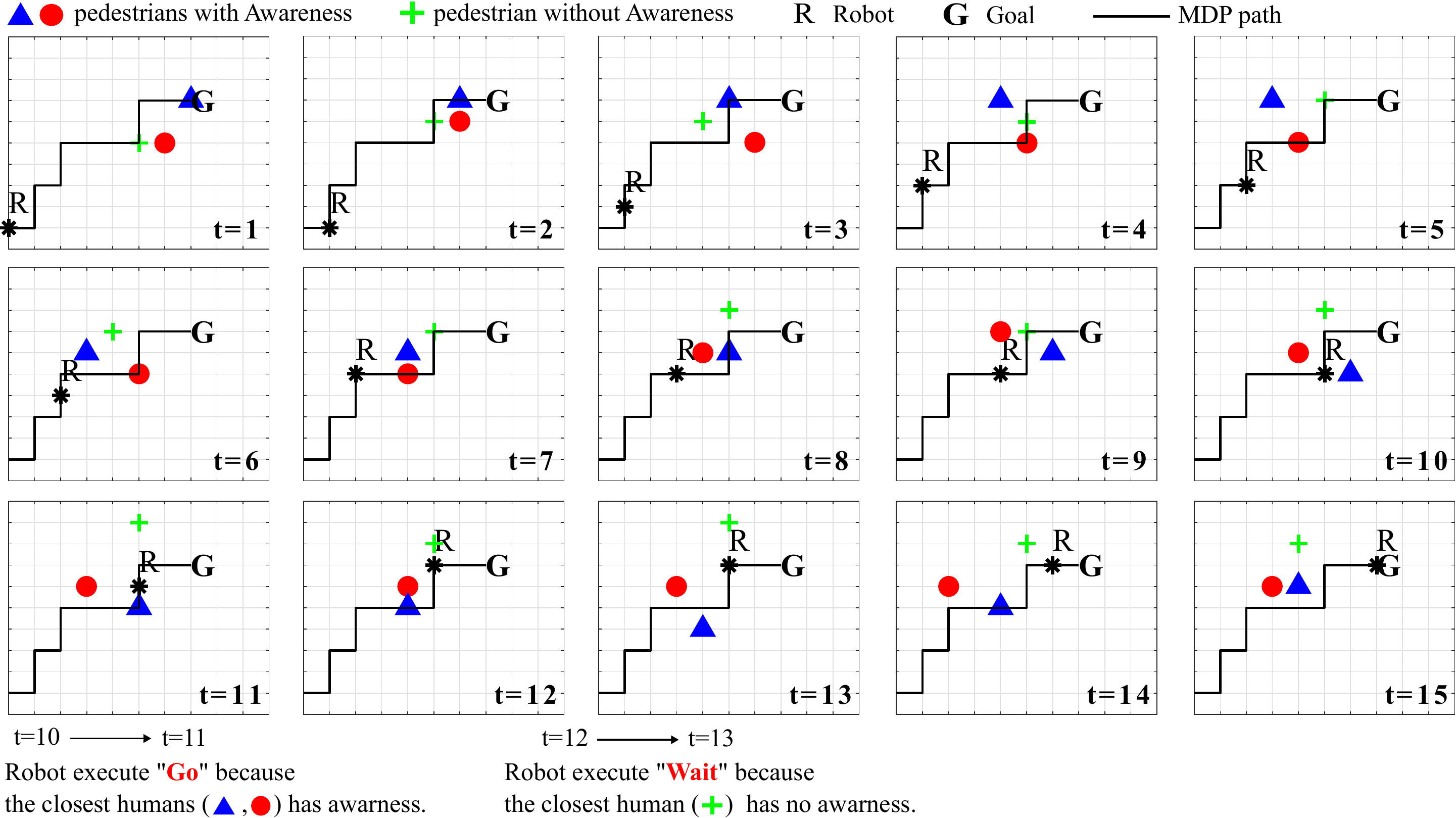}}
       \caption{Time series of grid map of local window (10 $\times$ 10). The robot goes on its own pre-calculated MDP path because the robot knows pedestrians who are aware of it. Otherwise, a robot executes "Wait" action if a pedestrian who is not aware of robot is nearby. }
       \label{Figure:Simulation result}
\end{figure*}

  \subsection{Belief Update} \label{Section:Belief Update}
  For every time step, POMDP maintains distribution over the states, or belief space. This belief space can be updated with the following equation:
    \begin{equation}
	b^{'}(s^{'}) = \eta Z(s^{'},a,o)\sum_{s \in S} T(s,a,s^{'})
 \end{equation} 
  
  where $\eta$ is a normalizing factor. DESPOT does not obtain the exact belief space, but calculates an approximate belief by a set of $K$ particles. Each particle corresponds to a sampled POMDP state, which can be written as, 
   \begin{align}
	B_t &:= (s^1_t,s^2_t, \cdots , s^K_t) \\
    s^i_{t+1} &= p(s_{t+1}|s_t,a_{t+1},o_{t+1})
 \end{align} 
	where each state vector, $s^i_t$ represents the state vector for the $i$th particle. A general particle filter is applied to update belief space with k, the number of particles, equal to 5000. This filtering is an approximation of Bayes filter approach. 

\subsection{Navigation Control}
  For basic navigation functions, ROS navigation stack  is utilized. When the global path planner finds the desired trajectory, it can generate safety velocity commands to control the wheels of mobile base to follow a desired trajectory according to information from odometry, sensor streams, and a goal pose.
  
\section{Experimental Results}
\subsection{System Description}
 A Toyota Human Support Robot (HSR) mobile robot is used as the hardware platform for both simulation and experiments. The mobile base consists of two drive omni-wheels and three casters, which are located to the front and rear of those. It can smoothly change the direction of navigation and has the capability to avoid obstacles. The maximum speed of HSR is approximately 0.22m/s, the maximum step size of the mobile base is 5mm, and the maximum incline is $5^{\circ}$. HSR has a variety of sensors. For vision information, two stereo cameras are mounted around the two eyes, a wide angle camera is on the forehead; a depth camera (Xtion, Asus) is placed on the top of the head to get RGB-D video stream. Furthermore, a laser range scanner (UST-20LX, Hokuyo) is equipped at the front side of the mobile base platform. The robot uses two different computers; one is for the main operating programs regarding basic functions and the other, which has a GPU (NVIDIA Jetson TK1), is only for running YOLO for object detection. All sub-programs to operate the robot are able to communicate, and to send and receive useful information to each other via the ROS interface.

\subsection{Pedestrian Model (Human)}
 One important feature of simulation is how to make movements in a pedestrian model. In our simulation, some pedestrians have their own path, others move randomly to collision-free space. They can even go out of the grid map, and the robot does not take them in to consideration anymore. The key feature of a pedestrian, the awareness variable, $G$, is determined by the gaze-detection. Although there are the limitations of continuously tracking people's faces in real situation, it is assumed that a camera has been able to keep track of the people's faces so it can track the movement of the pupil. During simulation, the awareness variables are manually set to each pedestrian. In contrast, for the real experiment, we assumed that once $G$ is activated, it never turns off so that the robot regards the person as being aware of it.  
 
\subsection{Scenario Analysis}
 We conducted both simulation and real platform experiments. For an actual experiment, there was a limit to closely analyzing all the situations. The evaluation of the algorithm is focused with the analysis result of the simulation.
 Each scenario has its own start and goal position. To test a robot's navigation performance according to the movements of pedestrians, we fixed the initial position and repeated the scenarios. The initial and goal positions of the robots are set to (0,1) and (7,7), respectively. For each scenario, 3 or 4 pedestrians are set to roam within the grid space. The initial positions of pedestrians are manually set to near the goal position. One pedestrian moves along the predefined path and others walk around with random direction in a local window. Each pedestrian has their mental state, $G$, the awareness variable, which is randomly chosen as a constant for each scenario. 
 
Fig. \ref{Figure:Simulation result} shows the one case scenario which contains the desired navigation strategy. A Robot is supposed to follow a global path (\bm$P$) and the DESPOT solver generates the optimal action policy depending on the belief states of the robot. In this scenario, it is assumed that pedestrians 1 and 2 (represented within blue triangle and red circle in Fig. \ref{Figure:Simulation result}) are aware of the robot and that pedestrian 3 (green cross) has no awareness of robot. Looking closely, at $t=7$ and $t=10$, a robot executed the 'Go' action because pedestrians who were aware are in front of it. This is because the robot assumes that they will not move in the direction where they will collide with it. At $t=12$, robot chose the 'Wait' action when the pedestrian who was not aware of it is near its predefined path. At the next time step, $t=13$, we confirmed that the robot moved along the path when the location of the untrustworthy pedestrian is a certain distance away. This scenario is a good example, showing that the robot changed its navigation policy according to how people are aware of a robot.  
 
 \subsection{Performance Analysis}
The proposed algorithm was tested in simulation with twenty-five sampled scenarios in simulation that have the same initial condition to ensure reliability. 

\subsubsection{Navigation time}
\begin{figure}[h]
    \centering
        {\includegraphics[width=0.95\linewidth]{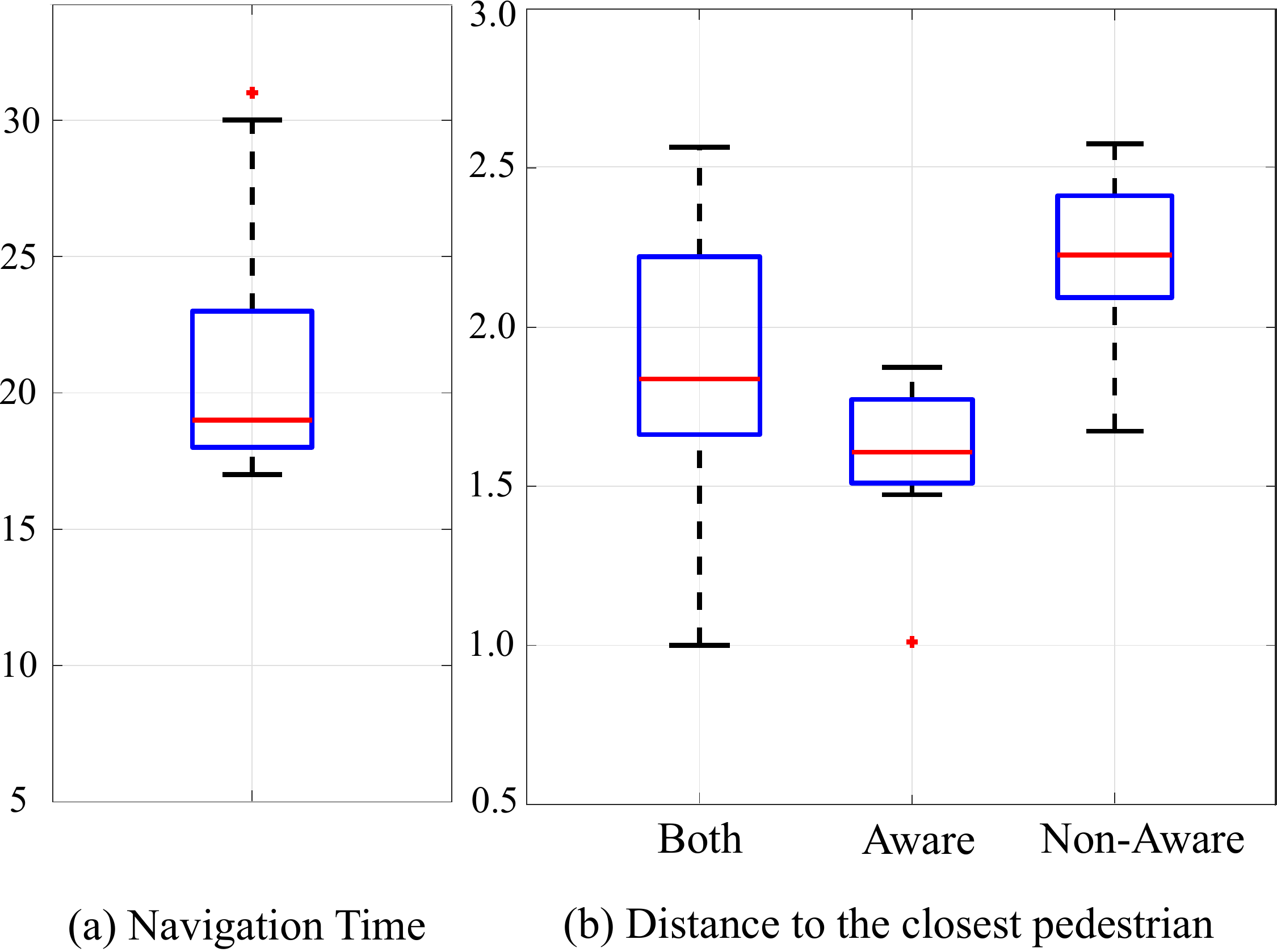}}
       \caption{Navigation performance : (a) Navigation Time (b) The distance between the closest human and robot for each cases. In both cases, there is one pedestrian with awareness, pedestrian without awareness.}
       \label{Figure:Simulation result2}
\end{figure}

 Due to a randomness of movement by pedestrians and the corresponding reactions of a robot, navigation time is different for each scenario. Since navigation time is also an important factor, we analyzed average navigation time to evaluate performance. If a robot never stops and keeps going along the path, a robot can reach the goal position point through 14 movements. As shown in Fig. \ref{Figure:Simulation result2}.(a), average navigation time (the number of steps to reach the goal position) is 20.54 and standard deviation equals to 3.97. When considering that there are 4 peoples, this time seems to be a reasonable navigation time.
 
\subsubsection{Proximity to pedestrians}
  The evaluation of navigation performance is quite difficult to define. To measure the social efficiency of the proposed navigation methods, we analyzed an average distance to the pedestrian when the robot executes the "Wait" action. By calculating this distance, we can examine how robot reacts to close pedestrians based on awareness. In Fig.\ref{Figure:Simulation result2}.(b), we first measured the mean distance from the closest pedestrian. The distance data is divided into two categories based on the existence of awareness. The two average distances between the pedestrian and robot, for the aware and non-aware cases were equal to 1.64 and 2.25, respectively. We confirmed that the robot achieves low proximity to a person who is not aware of the robot while it obtain high proximity to those who are aware of it. In other words, if pedestrians make eye contact with the robot during navigation, the robot can trust and navigate closer to them. As a result, by acquiring this characteristic of a distinctive proximity to humans, the social navigation ability of the robot improved.    

\section{CONCLUDING REMARKS}
The awareness-based navigation planning method is proposed for coexistence of humans and mobile robots. The proposed motion planner utilizes the concept of awareness as a simplified version of a human mental state. Both MDP and POMDP planners are integrated to increase social navigation performance while reducing real-time computational cost. To achieve the social interaction ability of a robot, a human detection and tracking system which includes a gaze detection model is implemented to obtain human positions and human awareness of the robot, which can be a key factor for socially-acceptable navigation. Adopting the concept of awareness, robots can react to or handle dynamic situations in a social manner, which is a key characteristic of human navigation. The simulation results and actual experiments with the HSR robot showed that the proposed planner makes it possible for a robot to handle ambiguous situations flexibly. If a person is aware of a robot, the robot is allowed to approach closer than to those who do not make eye contact, which indicates that they are not aware of the robot. 

 However, several future works still remain; currently, our proposed on-line POMDP planner currently can only select actions from discrete set. A next step is to extend this study to apply continuous-action sets. In addition, there exist hardware-implementation issues for mobile robots to detect awareness or other mental states. Human intention recognition technologies such as gaze-tracking or facial expression detection should be upgraded to estimate human mental state accurately. Lastly, an actual navigation of a robot should be evaluated properly from a human subject perspective. In other words, a metric to measure social-acceptableness for navigation need to be defined. Since performance of social navigation is quite difficult to measure, conducting human-related experiments and analyzing feedback from pedestrians can be a good reference for building desired strategies for a motion planner.




\section*{ACKNOWLEDGMENT}
This research was partially supported by a NASA Space Technology Research Fellowship (NSTRF), Grant number NNX15AQ42H.

\bibliographystyle{IEEEtran}
\bibliography{pomdp}

\begin{thebibliography}{10}
\providecommand{\url}[1]{#1}
\csname url@samestyle\endcsname
\providecommand{\newblock}{\relax}
\providecommand{\bibinfo}[2]{#2}
\providecommand{\BIBentrySTDinterwordspacing}{\spaceskip=0pt\relax}
\providecommand{\BIBentryALTinterwordstretchfactor}{4}
\providecommand{\BIBentryALTinterwordspacing}{\spaceskip=\fontdimen2\font plus
\BIBentryALTinterwordstretchfactor\fontdimen3\font minus
  \fontdimen4\font\relax}
\providecommand{\BIBforeignlanguage}[2]{{%
\expandafter\ifx\csname l@#1\endcsname\relax
\typeout{** WARNING: IEEEtran.bst: No hyphenation pattern has been}%
\typeout{** loaded for the language `#1'. Using the pattern for}%
\typeout{** the default language instead.}%
\else
\language=\csname l@#1\endcsname
\fi
#2}}
\providecommand{\BIBdecl}{\relax}
\BIBdecl

\bibitem{sisbot2007human}
E.~A. Sisbot, L.~F. Marin-Urias, R.~Alami, and T.~Simeon, ``A human aware
  mobile robot motion planner,'' \emph{IEEE Transactions on Robotics}, vol.~23,
  no.~5, pp. 874--883, 2007.

\bibitem{nonaka2004evaluation}
S.~Nonaka, K.~Inoue, T.~Arai, and Y.~Mae, ``Evaluation of human sense of
  security for coexisting robots using virtual reality. 1st report: evaluation
  of pick and place motion of humanoid robots,'' in \emph{Robotics and
  Automation, 2004. Proceedings. ICRA'04. 2004 IEEE International Conference
  on}, vol.~3.\hskip 1em plus 0.5em minus 0.4em\relax IEEE, 2004, pp.
  2770--2775.

\bibitem{shi2008human}
D.~Shi, E.~G. Collins~Jr, B.~Goldiez, A.~Donate, X.~Liu, and D.~Dunlap,
  ``Human-aware robot motion planning with velocity constraints,'' in
  \emph{Collaborative Technologies and Systems, 2008. CTS 2008. International
  Symposium on}.\hskip 1em plus 0.5em minus 0.4em\relax IEEE, 2008, pp.
  490--497.

\bibitem{mumm2011human}
J.~Mumm and B.~Mutlu, ``Human-robot proxemics: physical and psychological
  distancing in human-robot interaction,'' in \emph{Proceedings of the 6th
  international conference on Human-robot interaction}.\hskip 1em plus 0.5em
  minus 0.4em\relax ACM, 2011, pp. 331--338.

\bibitem{rios2015proxemics}
J.~Rios-Martinez, A.~Spalanzani, and C.~Laugier, ``From proxemics theory to
  socially-aware navigation: A survey,'' \emph{International Journal of Social
  Robotics}, vol.~7, no.~2, pp. 137--153, 2015.

\bibitem{kruse2013human}
T.~Kruse, A.~K. Pandey, R.~Alami, and A.~Kirsch, ``Human-aware robot
  navigation: A survey,'' \emph{Robotics and Autonomous Systems}, vol.~61,
  no.~12, pp. 1726--1743, 2013.

\bibitem{mainprice2011planning}
J.~Mainprice, E.~A. Sisbot, L.~Jaillet, J.~Cort{\'e}s, R.~Alami, and
  T.~Sim{\'e}on, ``Planning human-aware motions using a sampling-based costmap
  planner,'' in \emph{Robotics and Automation (ICRA), 2011 IEEE International
  Conference on}.\hskip 1em plus 0.5em minus 0.4em\relax IEEE, 2011, pp.
  5012--5017.

\bibitem{chernova2009interactive}
S.~Chernova and M.~Veloso, ``Interactive policy learning through
  confidence-based autonomy,'' \emph{Journal of Artificial Intelligence
  Research}, vol.~34, no.~1, p.~1, 2009.

\bibitem{helbing1995social}
D.~Helbing and P.~Molnar, ``Social force model for pedestrian dynamics,''
  \emph{Physical review E}, vol.~51, no.~5, p. 4282, 1995.

\bibitem{scovanner2009learning}
P.~Scovanner and M.~F. Tappen, ``Learning pedestrian dynamics from the real
  world,'' in \emph{Computer Vision, 2009 IEEE 12th International Conference
  on}.\hskip 1em plus 0.5em minus 0.4em\relax IEEE, 2009, pp. 381--388.

\bibitem{keller2014will}
C.~G. Keller and D.~M. Gavrila, ``Will the pedestrian cross? a study on
  pedestrian path prediction,'' \emph{IEEE Transactions on Intelligent
  Transportation Systems}, vol.~15, no.~2, pp. 494--506, 2014.

\bibitem{macrae2002you}
C.~N. Macrae, B.~M. Hood, A.~B. Milne, A.~C. Rowe, and M.~F. Mason, ``Are you
  looking at me? eye gaze and person perception,'' \emph{Psychological
  Science}, vol.~13, no.~5, pp. 460--464, 2002.

\bibitem{baker2014modeling}
C.~L. Baker and J.~B. Tenenbaum, ``Modeling human plan recognition using
  bayesian theory of mind,'' \emph{Plan, activity, and intent recognition:
  Theory and practice}, pp. 177--204, 2014.

\bibitem{hollands2002look}
M.~A. Hollands, A.~E. Patla, and J.~N. Vickers, ``“look where you’re
  going!”: gaze behaviour associated with maintaining and changing the
  direction of locomotion,'' \emph{Experimental brain research}, vol. 143,
  no.~2, pp. 221--230, 2002.

\bibitem{foka2007real}
A.~Foka and P.~Trahanias, ``Real-time hierarchical pomdps for autonomous robot
  navigation,'' \emph{Robotics and Autonomous Systems}, vol.~55, no.~7, pp.
  561--571, 2007.

\bibitem{pineau2003point}
J.~Pineau, G.~Gordon, S.~Thrun \emph{et~al.}, ``Point-based value iteration: An
  anytime algorithm for pomdps,'' in \emph{IJCAI}, vol.~3, 2003, pp.
  1025--1032.

\bibitem{silver2010monte}
D.~Silver and J.~Veness, ``Monte-carlo planning in large pomdps,'' in
  \emph{Advances in neural information processing systems}, 2010, pp.
  2164--2172.

\bibitem{ross2008online}
S.~Ross, J.~Pineau, S.~Paquet, and B.~Chaib-Draa, ``Online planning algorithms
  for pomdps,'' \emph{Journal of Artificial Intelligence Research}, vol.~32,
  pp. 663--704, 2008.

\bibitem{somani2013despot}
A.~Somani, N.~Ye, D.~Hsu, and W.~S. Lee, ``Despot: Online pomdp planning with
  regularization,'' in \emph{Advances in neural information processing
  systems}, 2013, pp. 1772--1780.

\bibitem{elfes1989using}
A.~Elfes, ``Using occupancy grids for mobile robot perception and navigation,''
  \emph{Computer}, vol.~22, no.~6, pp. 46--57, 1989.

\bibitem{redmon2016you}
J.~Redmon, S.~Divvala, R.~Girshick, and A.~Farhadi, ``You only look once:
  Unified, real-time object detection,'' in \emph{Proceedings of the IEEE
  Conference on Computer Vision and Pattern Recognition}, 2016, pp. 779--788.

\bibitem{bellotto2009multisensor}
N.~Bellotto and H.~Hu, ``Multisensor-based human detection and tracking for
  mobile service robots,'' \emph{IEEE Transactions on Systems, Man, and
  Cybernetics, Part B (Cybernetics)}, vol.~39, no.~1, pp. 167--181, 2009.

\bibitem{julier1997new}
S.~J. Julier and J.~K. Uhlmann, ``A new extension of the kalman filter to
  nonlinear systems,'' in \emph{Int. symp. aerospace/defense sensing, simul.
  and controls}, vol.~3, no.~26.\hskip 1em plus 0.5em minus 0.4em\relax
  Orlando, FL, 1997, pp. 182--193.

\bibitem{leigh2015person}
A.~Leigh, J.~Pineau, N.~Olmedo, and H.~Zhang, ``Person tracking and following
  with 2d laser scanners,'' in \emph{Robotics and Automation (ICRA), 2015 IEEE
  International Conference on}.\hskip 1em plus 0.5em minus 0.4em\relax IEEE,
  2015, pp. 726--733.

\bibitem{ageitgey2013}
\BIBentryALTinterwordspacing
A.~Geitgey. Face recognition package. [Online]. Available:
  \url{https://github.com/ageitgey/face_recognition}
\BIBentrySTDinterwordspacing

\bibitem{dalal2005histograms}
N.~Dalal and B.~Triggs, ``Histograms of oriented gradients for human
  detection,'' in \emph{Computer Vision and Pattern Recognition, 2005. CVPR
  2005. IEEE Computer Society Conference on}, vol.~1.\hskip 1em plus 0.5em
  minus 0.4em\relax IEEE, 2005, pp. 886--893.

\bibitem{kazemi2014one}
V.~Kazemi and J.~Sullivan, ``One millisecond face alignment with an ensemble of
  regression trees,'' in \emph{Proceedings of the IEEE Conference on Computer
  Vision and Pattern Recognition}, 2014, pp. 1867--1874.

\bibitem{timm2011accurate}
F.~Timm and E.~Barth, ``Accurate eye centre localisation by means of
  gradients.'' \emph{VISAPP}, vol.~11, pp. 125--130, 2011.

\bibitem{ge2000new}
S.~S. Ge and Y.~J. Cui, ``New potential functions for mobile robot path
  planning,'' \emph{IEEE Transactions on robotics and automation}, vol.~16,
  no.~5, pp. 615--620, 2000.

\end{thebibliography}

\end{document}